\pdfoutput=1
\documentclass[11pt]{article}

\usepackage{acl}

\usepackage{times}
\usepackage{latexsym}

\usepackage[T1]{fontenc}
\usepackage[utf8]{inputenc}

\usepackage{microtype}

\usepackage{graphicx}
\usepackage{multirow} 
\usepackage{multicol}
\usepackage{makecell}
\usepackage{tabularx}
\usepackage{color}
\usepackage{xcolor}
\usepackage{listings}
\usepackage{subcaption}
\usepackage{url}
\usepackage{latexsym}
\usepackage{booktabs}
\usepackage{adjustbox}
\usepackage{url}
\usepackage{xspace}

\newcommand{\ignore}[1]{}

\newcommand{\uniqa}{\texttt{UniK-QA}\xspace}
\newcommand{\uniqat}{UniK-QA\xspace}

\newcommand*\samethanks[1][\value{footnote}]{\footnotemark[#1]}
\newcommand{\secref}[1]{\S\ref{#1}}

\title{\uniqat: Unified Representations of Structured and Unstructured Knowledge for Open-Domain Question Answering}

\author{Barlas Oğuz\textsuperscript{1}\thanks{\hspace{.06in}Equal contribution}, Xilun Chen\textsuperscript{1}\samethanks, Vladimir Karpukhin\textsuperscript{1},\\ \textbf{Stan Peshterliev\textsuperscript{1}, Dmytro Okhonko\textsuperscript{1}, Michael Schlichtkrull\textsuperscript{2,3}\thanks{\hspace{.06in}Work done while interning with Meta AI.},} \\ \textbf{Sonal Gupta\textsuperscript{1},  Yashar Mehdad\textsuperscript{1}, Wen-tau Yih\textsuperscript{1}} \\
\textsuperscript{1}Meta AI,
\textsuperscript{2}University of Amsterdam,
\textsuperscript{3}University of Edinburgh,\\
\scalebox{0.9}{{\tt \{barlaso,xilun,stanvp,sonalgupta,mehdad,scottyih\}@fb.com}}\\
\scalebox{0.87}{{\tt vlad.karpuhin@gmail.com, d.okhonko@gmail.com, m.s.schlichtkrull@uva.nl}}}

\date{}

\begin{document}
\maketitle
\begin{abstract}
We study open-domain question answering with \emph{structured, unstructured} and \emph{semi-structured} knowledge sources, including text, tables, lists and knowledge bases.  
Departing from prior work, we propose a unifying approach that homogenizes all sources by reducing them to text and applies the retriever-reader model which has so far been limited to text sources only.
Our approach greatly improves the results on knowledge-base QA tasks by 11 points, compared to latest graph-based methods.
More importantly, we demonstrate that our \emph{unified knowledge} (\uniqa{}\footnote{The code of \uniqa{} is available at: \url{https://github.com/facebookresearch/UniK-QA}.}) model is a simple and yet effective way to combine heterogeneous sources of knowledge, advancing the state-of-the-art results on two popular question answering benchmarks, NaturalQuestions and WebQuestions, by 3.5 and 2.6 points, respectively.

\end{abstract}

% Sections
\section{Introduction}

\begin{figure}[t!]
\includegraphics[width=\linewidth]{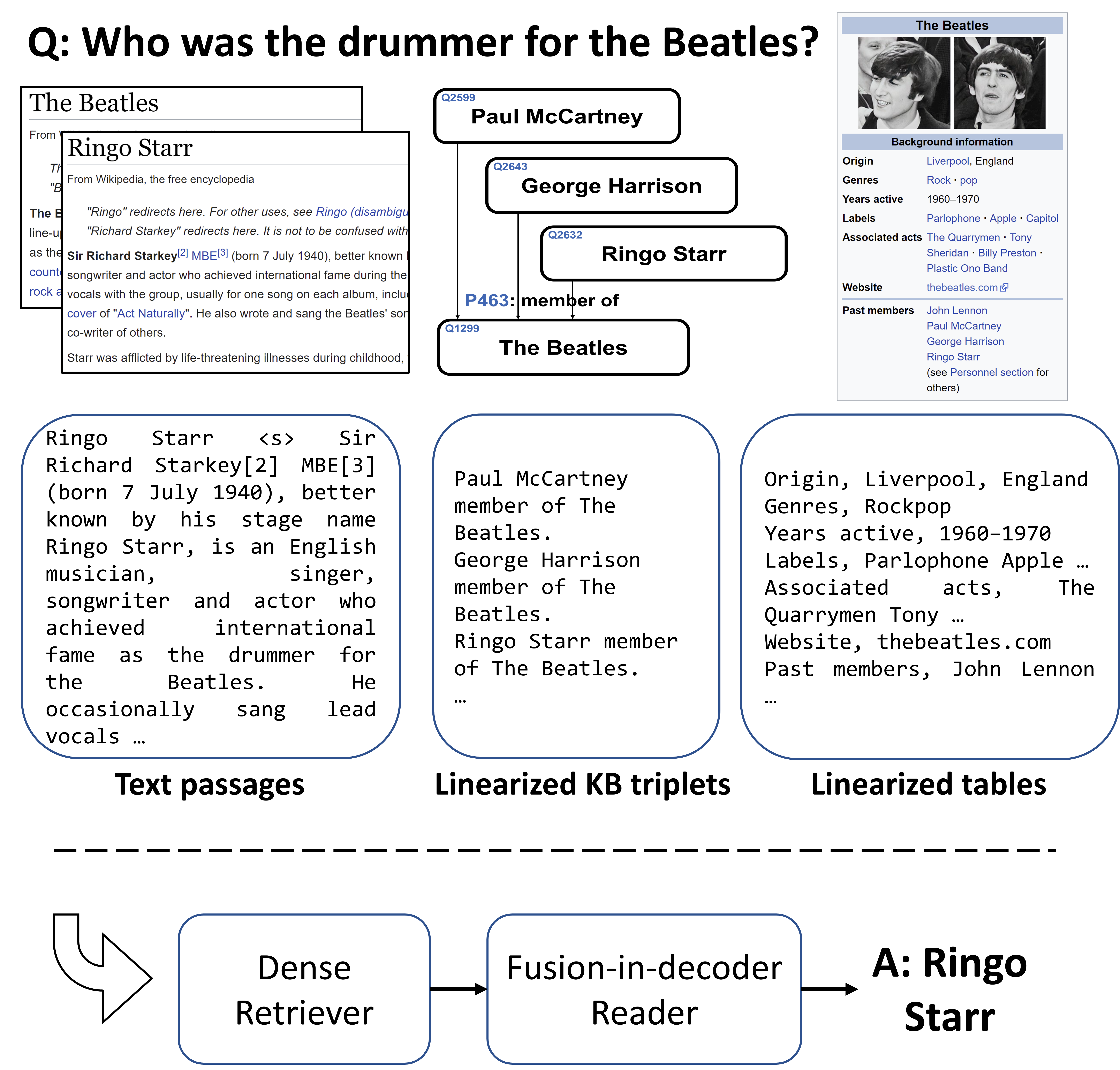}
\caption{Illustration of \uniqa{}'s workflow for unified-knowledge question answering: Heterogeneous information sources are linearized into text. A dense retriever retrieves passages from a mix of sources, which are jointly processed by the reader to produce the answer.}
\centering
\label{fig:hybridqa}
\vspace{-3mm}
\end{figure}

Answering factual questions has long been an inspirational challenge to information retrieval and artificial intelligence researchers~\cite{voorhees-tice-2000-trec,lopez2011question}.
In its most general form, users can ask about \emph{any} topic and the answer may be found in \emph{any} information source.  
Defined as such, the challenge of \emph{open domain question answering} is extremely broad and complex.
Though there have been successful undertakings which embrace this complexity~\citep[notably][]{ferrucci2012introduction}, most recent works make simplifying assumptions as to the source of answers, which
fall largely in two categories: \textbf{structured data} and \textbf{unstructured text}.  

A long line of research aims to answer user questions using a structured \emph{knowledge base}~(KB)~\citep{WebQ, yih2015semantic}, known as \textbf{KBQA}.
Typically, a KB can be viewed as a knowledge graph consisting of entities, properties, and a pre-defined set of relations between them. 
A question can be answered, provided that it can be expressed within the language of relations and objects present in the knowledge graph.
With a high-quality, carefully curated KB, answers can be extracted with fairly high precision.
KBQA, however, struggles with low answer coverage due to the cost of curating an extensive KB, as well as the fact that many questions simply cannot be answered using a KB if the answers are not entities.

A second line of work targets a large collection of unstructured text~(such as Wikipedia)~\citep{drqa} as the source of answers.
Thanks to the latest advances in machine reading comprehension and text retrieval, substantial progress has been made for open-domain question answering from text (\textbf{TextQA}) in just the past couple years~\citep{yang2019end, ORQA, DPR, REALM, izacard-grave-2021-leveraging}.
On the other hand, semi-structured tables and structured KBs can be valuable knowledge sources, yet TextQA methods are restricted in taking only unstructured text as input, missing the opportunity of using these complementary sources of information to answer more questions.

When it comes to answering questions using both structured and unstructured information, a straightforward solution is combining specialized TextQA and KBQA systems.
The input question is sent to multiple sub-systems, and one of them is selected to output the final answer.
While this approach may take advantage of the state-of-the-art models designed for different information sources, the whole end-to-end system becomes fairly complex. It is also difficult to handle questions that require reasoning with information from multiple sources.

Having a more integrated system design that covers heterogeneous information sources has proven to be difficult. 
One main reason is that techniques used for KBQA and TextQA are drastically different.
The former exploits the graph structure and/or semantic parsing to convert the question into a structured query, while TextQA has mostly settled on the retriever-reader architecture powered by pre-trained transformers.
Recent work on multi-source QA has tried to incorporate free text into graph nodes~\citep{sun2018open, lu2019answering} to make texts amenable to KBQA methods, but the performance remains unconvincing. 

In this work, we propose a novel \emph{unified knowledge representation} (\uniqa{}) approach for open-domain question answering with heterogeneous information sources.
Instead of having multiple specialized sub-systems or incorporating text into knowledge graphs, we \textit{flatten} the structured data and apply TextQA methods.  Our main motivation for doing so is to make the powerful machinary of pre-trained transformers available for structured QA. 
In addition, this approach opens the door to a simple and unified architecture.  We can easily support semi-structured sources such as lists and tables, as well as fully structured knowledge bases.
Moreover, there is no need to specially handle the schema or ontology that defines the structure of the KB, making it straightforward to support multiple KBs.
Our \uniqa{} model incorporates some 27 million passages composed of text and lists, 455,907 Wikipedia tables, and 3 billion relations from two knowledge bases (Freebase and Wikidata) in a single, unified open-domain QA model.

We first validate our approach by modeling KBQA as a pure TextQA task.  We represent all relations in the KB with their textual surface form, and train a \emph{retriever-reader} model on them as if they were text documents.  This simple approach works incredibly well, improving the exact match score on the WebQSP dataset by $11\%$ over previous state of the art. 
This result further justifies our choice of unifying multi-source QA under the TextQA framework as it can improve KBQA performance \emph{per se}.

For our multi-source QA experiments, we consider lists, tables, and knowledge bases as sources of structured information, and convert each of them to text using simple heuristics.  We model various combinations of structured sources with text, and evaluate on four popular open-domain QA datasets, ranging from entity-heavy KBQA benchmarks to those targeting free-form text sources.
Our results indicate that our multi-source \uniqa{} approach, unlike existing efforts on combining KBQA and TextQA, consistently improves over strong TextQA baselines in all cases.
We obtain new state-of-the-art results for two datasets, advancing the published art on NaturalQuestions by 3.5 points and on WebQuestions by 2.6 points.

In addition, we consider the realistic setting in which the source of questions is not known \emph{a priori}, as would be the case for a practical system. 
We train a single \emph{multi-dataset} model on a combined dataset from several benchmarks, and show that it outperforms all single-source baselines across this diverse set of questions.

\section{Background \& Related Work}

\subsection{Knowledge-base question answering (KBQA)}

A knowledge base (KB) considered in this work is a collection of facts, represented as a set of subject-predicate-object \emph{triples}.
Each triple $(e_1, p, e_2)$ denotes a binary relationship between the subject entity $e_1$ and the object $e_2$ (e.g., places, persons, dates or numbers), as well as their relation type, or predicate $p$ (e.g., \emph{capital\_of}, \emph{married\_to}, etc.).

Modern large-scale KBs, such as Freebase~\citep{bollacker2008freebase}, DBPedia~\citep{auer2007dbpedia} and Wikidata~\citep{vrandevcic2014wikidata} can contain billions of triples that describe relations between millions of entities, 
making them great sources of answers to open-domain questions.
The prevailing approach for knowledge-base question answering (KBQA) is semantic parsing~\citep{WebQ, yih2015semantic}, where a natural language question is converted into a logical form that can be used to query the knowledge base.
Such methods are tailored to the specific graph structure of the KB and are usually not directly applicable to other knowledge sources.

\subsection{Open-domain question answering from text (TextQA)}
KBQA is ultimately limited in its coverage of facts and the types of questions it can answer.  On the other hand, large collections of text such as Wikipedia or CommonCrawl promise to be a richer source of knowledge for truly open domain question answering systems.  This line of work (which we will refer to as TextQA) has been popularized by the TREC QA tracks \citep{voorhees-tice-2000-trec}, and has seen explosive growth with the advent of neural machine reading (MRC) \citep{rajpurkar-etal-2016-squad} models.  In the neural era, \citet{drqa} were the first to combine MRC with retrieval for end-to-end QA.  Subsequent work cemented this \emph{retriever-reader} paradigm, with improved reader models \citep{yang2019end, izacard-grave-2021-leveraging} and neural retrievers \citep{ORQA, REALM, DPR}.  Despite impressive advances, TextQA systems still underperform KBQA, especially on benchmarks originally created for KBs such as WebQuestions.  
Furthermore, they also fall short of universal coverage, due to the exclusion of other (semi-)structured information sources such as tables.

\subsection{Question answering from tables}
Large amounts of authoritative data such as national statistics are often available in the form of tables.
While \mbox{KBQA} and \mbox{TextQA} have enjoyed increasing popularity, tables as a source of information has surprisingly escaped the attention of the community save for a few recent works. 

Working with web tables can be challenging, due to the lack of formal schema, inconsistent formatting and ambiguous cell values (e.g., entity names).  In contrast to relational databases and KBs, tables can at best be described as \emph{semi-structured} information.  \citet{sun2016table} considered open domain QA from web tables, however made no use of unstructured text.  Some recent work investigated MRC with tables without a retrieval component \citep{pasupat-liang-2015-compositional,yin2020tabert, chen2019tabfact}.  In addition, \citet{chen2020open, chen2020hybridqa} investigated open domain QA using tables and text.
While they are in a similar direction, these works focus on complex, crowd-sourced questions requiring more specialized methods, while we target the case of simple, natural questions and investigate if popular TextQA and KBQA benchmarks can be further improved with the addition of tables.

\subsection{Fusion of text and knowledge-base}
As discussed, KBQA and TextQA are intuitively complementary, and several attempts have been made to merge them to get the benefits of both.  An early example is \cite{ferrucci2012introduction}, which combines multiple expert systems and re-ranks them to produce the answer.  More recent work attempts to enrich the KB by extracting structure from text.  One way to accomplish this is using \mbox{OpenIE} triplets \citep{fader2014open, xu2016hybrid}, thus staying completely within the semantic parsing paradigm.  Somewhat closer to our approach are UniversalSchemas \citep{riedel2013relation, das2017question}, which embed KB relations and textual relations in a common space.  Yet, UniversalSchemas are also constrained to an entity-relation structure.  The latest in this line are the works of \cite{sun2018open, sun2019pullnet}, which augments the knowledge graph with text nodes and applies graph methods to identify candidate answers.

By retaining structure, previous work was able to take advantage of KBQA methods, but also failed to capture the full richness of TextQA.  We depart radically in our approach, by foregoing all structure, and directly applying TextQA methods based on the more general \emph{retriever-reader} architecture.  We also evaluate on a more diverse benchmark set composed of natural open domain datasets, as well as those originally meant for KBQA, and demonstrate strong improvements in this truly open-domain setting.
Concurrent work~\citep{agarwal-etal-2021-knowledge} proposed a similar idea for language model pre-training and also evaluated on open-domain QA.  Our work differs in that (1)~we have a more comprehensive treatment of sources (including tables, lists and multiple KBs) and ODQA datasets, (2)~we compare against and improve on much stronger state-of-the-art baselines, and (3)~we also evaluate in a more realistic multi-dataset setting with all datasets handled by a single model.

\section{Modeling}

\subsection{\uniqa{} architecture}
We use a retriever-reader architecture, with \emph{dense passage retriever} (DPR) \citep{DPR} as retriever and \emph{fusion-in-decoder} (FiD) \citep{izacard-grave-2021-leveraging} as our reader.  
Structured knowledge such as tables, lists and KB relations are converted to text with simple heuristics (\secref{sec:model:kbqa}, \secref{sec:model:tables}), and we generalize DPR to retrieve from these heterogeneous documents as well as regular text passages.
Each retrieved document is concatenated with the question, then independently encoded by the reader encoder.  Fusion of information happens in the decoder, which computes full attention over the entire concatenated input representations.  The overall architecture is illustrated in Figure~\ref{fig:hybridqa}.

\paragraph{Retriever}
The DPR retriever consists of a dense document encoder and a question encoder, trained such that positive documents have embeddings closer to the question embedding in dot product space.  We follow the original DPR implementation and hyperparameters (see~\secref{sec:implementation}).
We further include tables, lists and KB relations in the index.  The details of how these are processed into documents and merged are in the subsequent sections.

One improvement we make to the training process is iterative training, where better hard negatives are mined at each step using the model at the previous step, similar to \citep{xiong2020approximate}.  All models including our text-only baselines benefit from this change.  We find 2 iterations sufficient.

\paragraph{Reader}
The FiD reader has demonstrated strong performance in the text-only setting and effective in fusing information from a large number of documents~\citep{izacard-grave-2021-leveraging}.  We thus find it a natural candidate for fusing knowledge from various sources.
We use the FiD model with T5-large \citep{t5}, 100 context documents, and the original hyper-parameters for all experiments.
See~\secref{sec:implementation} for more implementation details.

\subsection{Unified representations for KBs}\label{sec:model:kbqa}

\begin{figure}
    \centering
    \includegraphics[width=\linewidth]{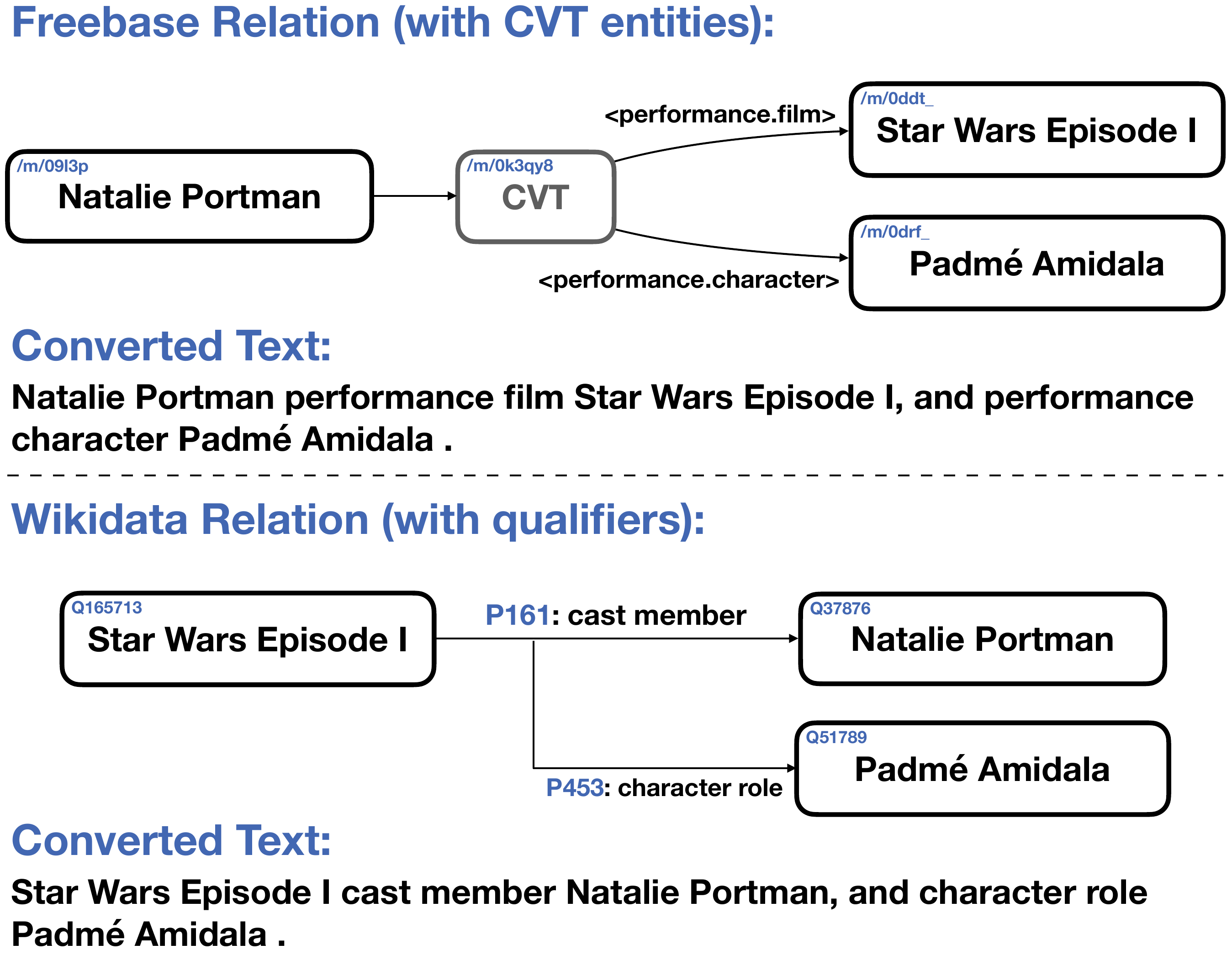}
    \caption{Converting Freebase and Wikidata relations to text.}
    \label{fig:kb_to_text_example}
    \vspace{-3mm}
\end{figure}

In order to apply our retriever-reader model, we first convert KB relations into text using simple heuristics.
For a relation triple $\langle subj, pred, obj\rangle$, where $subj$, $pred$ and $obj$ are the subject, predicate and object of the relation respectively, we serialize it by concatenating the text surface forms of $subj$, $pred$ and $obj$.

More complex ($n$-ary) relations involve multiple predicates and objects, such as \emph{Natalie Portman played the character Padm\'e Amidala in the movie Star Wars}, and can be expressed differently depending on the KB.
In particular, Freebase uses \emph{compound value types} (CVTs) to convert an $n$-ary relation into multiple standard triples, while Wikidata allows a predicate to have \emph{qualifiers} to express additional properties~\cite{freebase2wikidata}.
In this work, we convert an $n$-ary relation into a single sentence by forming a comma-separated clause for each predicate (Figure~\ref{fig:kb_to_text_example}).\footnote{A side benefit of this approach is that these complex relations are now represented as a single piece of text, whereas they would normally be considered multi-hop and require more complex methods \citep{fu2020survey} if using traditional graph-based KBQA models.}

Besides our heuristic-based linearization of KB relations, there are alternatives such as template-based or model-based methods.
Since KBs such as Freebase and Wikidata have hundreds of thousands of different types of relations, it is prohibitive to come up with templates for each relation type.
On the other hand, model-based linearization achieves worse retrieval recall than our simple heuristics despite being much more expensive.
In particular, we experiment with a top-ranked KB-to-text model~\cite{li-etal-2020-leveraging-large} from the WebNLG 2020 challenge~\cite{castro-ferreira-etal-2020-2020}, which is based on T5-large.
Preliminary results on KBQA show that the WebNLG model achieves a 87.9\% retrieval recall @100 on the dev set of WebQSP~\cite{yih2016the}, while our simple heuristics performs better at 94.7\%.
We hence stick with our simple heuristics in all experiments.

Once converted to text, relations can be indexed and retrieved using DPR.
We use existing TextQA DPR checkpoints for retrieving KB relations without any retraining.
We index individual relations to best leverage the power of DPR for retrieving the most relevant relations for a given question\footnote{Indexing at a coarser granularity (such as creating a document for each entity) also has practical challenges because certain entities (e.g.,~United States) may have hundreds of thousands of relations, resulting in extremely long documents.}.
Unlike most existing KBQA works, our approach can also seamlessly incorporate multiple KBs by storing all relations into a joint index and retrieving from it (see \secref{sec:exp:results}).

Directly indexing billions of relations in the entire KB can bring additional engineering challenges.  To avoid these, we implement retrieval of relations in two steps, where an entity linking system is used in the first step to narrow down the search to a high-recall 2-hop neighborhood of the retrieved entities for each question (We use STAGG~\citep{yih2015semantic} in the case of Freebase and ELQ \citep{li2020efficient} for Wikidata). 
We then use DPR to retrieve relations from this reduced set. 
As the relation representations are usually short sentences, we combine retrieved relations into passages of at most 100 tokens, after which they are fed to the FiD reader in the same way as text paragraphs.

\subsection{Unified representations for lists \& tables}\label{sec:model:tables}
\citet{DPR} excludes lists and tables from their passage collection.
For lists, we simply retain them as part of the text documents without special preprocessing, which improves retrieval recall in our experiments (see Table~\ref{tab:tables} in \secref{sec:analysis}).
We now discuss about our treatment of tables.

English Wikipedia contains more than 3 million tables (`classical' tables embedded in text as well as specialized tables like info-boxes), which are a huge source of factual knowledge by themselves and can substantially increase the coverage of open-domain QA systems.  
For instance, the answer to approximately a quarter of the questions in the NaturalQuestions (NQ) dataset can be found in Wikipedia tables~\cite{NQ}.

We start from a large subset of Wikipedia tables extracted and released as part of the NaturalQuestions dataset.  We include all candidate documents which are part of the training set, extract nested tables into independent units, and filter out single-row tables as well as `service' tables.  This results in a corpus of 455,907 tables, which are used in our experiments.

As with KB relations, semi-structured content in tables need to be `linearized' into text for the \emph{retriever-reader} model to work.  There are many ways to do such linearization~\citep[see][]{yin2020tabert,chen2019tabfact}.
We tried two types of tables linearization: `template'-like encoding used in recent literature~\cite{chen2019tabfact} and a simpler one which we find works the best in our experiments (see Table~\ref{tab:tables}, bottom half).
In particular, we concatenate cell values on the same row, separated by commas, to form the text representation, and multiple rows are then combined into longer documents delimited by newlines.
As with TextQA, we divide linearized tables into 100-token chunks for indexing and retrieval.  We take the first non-empty table row as the \emph{header} and include it in every table chunk.  This heuristic to select the first non-empty row as header is crucial and adds 4-6 points to top-20 passage accuracy.

\section{KBQA as TextQA: A Motivating Experiment}\label{sec:exp:kbqa_as_textqa}

In this section, we present a motivating experiment showing that our \uniqa{} approach not only provides a natural pathway to multi-source open-domain QA, but also improves KBQA per se.
In particular, we evaluate our approach on a widely-used KBQA dataset, WebQSP~\cite{yih2016the}, in the single-source setting.

\begin{table}[t]
\begin{center}
\begin{tabular}{lr}
\toprule
\textbf{Model} & \textbf{Hits@1} \\ 
\midrule
GraftNet \cite{sun2018open} & 67.8  \\
PullNet \cite{sun2019pullnet} & 68.1  \\ 
\emph{EmQL~\cite{sun2020faithful}} & \emph{75.5*} \\
\midrule
Our KBQA (T5-base) & 76.7  \\
Our KBQA (T5-large) & \textbf{79.1}  \\
\bottomrule
\end{tabular}
\end{center}
\caption{\label{webqsp} Hits@1 on WebQSP dataset using Freebase. (*)EmQL uses oracle entities, hence is not directly comparable with the others.}
\label{tab:webqsp}
\vspace{-3mm}
\end{table}

We use Freebase as the knowledge source, and re-use pre-computed STAGG entity linking results and 2-hop neighborhoods as provided by~\citet{sun2018open} for fair comparisons.  We convert KB relations in the 2-hop neighborhood into text, retrieve the most relevant ones using DPR to form 100 context passages, and feed them into the T5 FiD reader as described in Section~\ref{sec:model:kbqa}.
We use the original DPR checkpoint from \citet{DPR} for retrieval, and train FiD using the training questions in WebQSP and the DPR-retrieved contexts with default hyperparameters (see~\secref{sec:implementation}).
The results are shown in Table~\ref{tab:webqsp}, where the numbers represent \emph{Hits@1}, or the percentage of the model's top-predicted answer being a ``hit'' (exact match) against one of the gold-standard answers.

We see that our KBQA method outperforms previous state-of-the-art methods by a wide margin, improving exact match accuracy to $79.1\%$.  Since we adopt the exact same KB setup and pre-processing procedure from previous work, this improvement can be attributed purely to our \uniqa{} model.  We take this result as strong evidence for our claim that powerful TextQA methods generalize well to structured data, and offer a natural new framework for unifying structured and unstructured information sources.

\section{Multi-Source QA Experiments}\label{sec:experiments}

\begin{table*}[t]
\centering

\begin{tabular}{l@{\hskip 3em}ccccc}
 \toprule
 Model  & \textbf{NQ}  & \textbf{WebQ} & \textbf{Trivia} & \textbf{TREC} & \textbf{Avg.} \\
\midrule
SoTA & 51.4${}^1$ & 55.1${}^3$ & 67.6${}^1$ & 55.3${}^2$ & 57.3 \\
Retrieval-free & 28.5${}^4$ & 30.6${}^4$ & 28.7${}^4$ & - & - \\
\midrule
\multicolumn{1}{l}{\textit{Per-dataset models}} & \multicolumn{4}{l}{} \\
Text & 49.0 & 50.6 & 64.0 & 54.3 & 54.5  \\
Tables & 36.0 & 41.0 & 34.5 & 32.7 & 36.1  \\
KB & 27.9 & 55.6 & 35.4 & 32.4 & 37.8  \\
Text + tables & \textbf{54.1} & 50.2 & \textbf{65.1} & 53.9 & 55.8  \\
Text + tables + KB & 54.0 & \textbf{57.8} & 64.1 & \textbf{55.3} & \textbf{57.8}  \\
\midrule
\multicolumn{1}{l}{\textit{Multi-dataset model}} & \multicolumn{4}{l}{} \\
Text & 50.3 & 45.0 & 62.6 & 45.7 & 50.9  \\
Tables & 34.2 & 38.4 & 33.7 & 31.1 & 34.4  \\
KB & 25.9 & 43.3 & 34.2 & 38.0 & 35.4  \\
Text + tables & \textbf{54.6} & 44.3 & \textbf{64.0} & 48.7 & 52.9  \\
Text + tables + KB & 53.7 & \textbf{55.5} & 63.4 & \textbf{51.3} & \textbf{56.0}  \\
\bottomrule
\end{tabular}
\caption{Exact match results on the test set.  SoTA numbers are from \citep{izacard-grave-2021-leveraging}${}^1$, \citep{iyer-etal-2021-reconsider}${}^2$ which are TextQA approaches,  and \cite{jain-2016-question}${}^3$, which is a KBQA method. \cite{jain-2016-question} reports another metric; however, their predictions are available from which we calculated the EM score. Retrieval-free numbers refer to closed-book results from \citet{t5close}${}^4$ with the same T5 model.}
\label{tab:main}
\vspace{-3mm}
\end{table*}

\begin{table}[t]
%\small
\centering

\begin{tabular}{@{\hspace{0.3em}}lc@{\hspace{0.4em}}c@{\hspace{0.4em}}c@{\hspace{0.4em}}c@{\hspace{0.3em}}}
 \toprule
 Source(s)  & \textbf{NQ}  & \textbf{WebQ} & \textbf{Trivia} & \textbf{TREC}\\
\midrule
KB-only (1 KB) & 27.9 & 55.6 & 35.4 & 32.4  \\
KB-only (2 KBs) & 30.9 & 56.7 & 41.5 & 36.0 \\
All (1 KB) & 54.0 & \textbf{57.8} & 64.1 & \textbf{55.3}\\
All (2 KBs) & \textbf{54.9} & 57.7 & \textbf{65.5} & 54.0 \\
\bottomrule
\end{tabular}
\caption{Results for combining Freebase and Wikidata.}
\label{tab:kbc}
\vspace{-3mm}
\end{table}

We now present our main experiments on unified multi-source question answering.

\subsection{Datasets}
For our main experiments, we use the same datasets that have recently become somewhat standard for evaluating open-domain QA~\citep{ORQA}:

\noindent
\textbf{NaturalQuestions (NQ)}~\cite{NQ} consists of questions mined from real Google search queries and Wikipedia articles with answer spans annotated. While the answer spans are usually on the regular, free-form text, some span annotations are in tables.

\noindent
\textbf{WebQuestions (WebQ)}~\cite{WebQ} targets Freebase as the source of answers, with questions coming from Google Suggest API.

\noindent
\textbf{TriviaQA (Trivia)}~\cite{joshi-etal-2017-triviaqa} contains a set of trivia questions with answers originally scraped from the Web.

\noindent
\textbf{CuratedTREC (TREC)}~\cite{baudivs2015modeling} is a collection of questions from TREC QA tracks and various Web sources, intended to benchmark open-domain QA on unstructured text.

\subsection{Combinations of sources}\label{sec:exp:combination}
We compare 5 variations of our model, each with a different combination of information sources.  We have \emph{Text}-only, \emph{Tables}-only and \emph{KB}-only variants as single-source baselines.  Next, the \emph{Text + tables} model makes use of the entire Wikipedia dump, including lists and tables.  Finally we add the KBs resulting in the \emph{Text + tables + KB} model. 

The \emph{Text + tables} model uses a unified dense index, where text passages and table chunks are jointly indexed. 
For the \emph{Text + tables + KB} model, the KB relations are indexed separately.
As described in \secref{sec:model:kbqa}, we use DPR to retrieve individual KB relations for each question, and the top-scoring KB relations are concatenated into 100-token passages to be fed to the reader.
These passages are then merged with the passages retrieved from the \emph{Text + tables} index using a fixed quota for KB relations.
This quota is determined by maximizing retrieval recall on the development set (see~\secref{sec:implementation:merge}). 
We also experiment with combining multiple KBs by using DPR to jointly retrieve from all relations of both KBs, which is straightforward to implement with our approach despite differences in the KB structure.

\subsection{A multi-dataset model}\label{sec:exp:multidataset}

In a realistic setting, the best knowledge source to answer a given question is unknown \emph{a priori} to the system, but most open-domain QA datasets are collected with respect to a specific information source (e.g., Wikipedia for NQ and Freebase for WebQ).
To better simulate the real-world scenario, we also experiment with a setting where we train a single model on the combination of all 4 datasets and evaluate without any input to the model as to the source of questions.\footnote{We normalize the questions by removing question marks and by presenting them in lowercase.}  We refer to this as the \emph{multi-dataset} setting.  This setting was previously investigated in several works \citep{DPR, maillard-etal-2021-multi, qi-etal-2021-answering}, but not in the multi-source context. We train multi-dataset models for all 5 variants described above.  The smaller datasets, WebQ and TREC, are upsampled 5 and 8 times respectively while training.

\subsection{Results}\label{sec:exp:results}

Main results are presented in Table~\ref{tab:main}. 
In the first set of experiments, we train a reader model independently for each dataset, as typically done in previous work. We use Freebase as knowledge base for WebQuestions as intended, and use Wikidata for all others.  The multi-dataset model uses Wikidata.

The results highlight the limitation of current state-of-the-art open-domain QA models which use texts as the only information source.
On WebQ, for instance, the KB-only model performs 5\% better than the text-only one, and previous state of the art is also achieved by the KBQA model.
Moreover, adding structured information sources significantly improves the performance over text-only models on \emph{all} datasets, obtaining state-of-the-art results for NQ, WebQ and TREC.
This indicates that KBs and tables contain valuable knowledge which is either absent in the unstructured texts or harder to extract from them (see also \secref{sec:analysis}).

In the \emph{multi-dataset} setting, we also observe clear improvements from combining sources, with the \emph{Text + tables + KB} model outperforming the \emph{Text}-only baseline by 5.4 points on average.
The performance is generally lower than the per-dataset models, especially for the small datasets (WebQ and TREC), which may be due to the fact that each of these datasets was collected on a single information source and the multi-dataset model is less likely to exploit this prior knowledge.

\paragraph{Multiple KBs} 
We also experiment with combining \emph{both} Wikidata and Freebase.  We see substantial improvements on all datasets in the KB-only setting over using a single KB, as well as significant gains over our best numbers for NQ and TriviaQA in the \emph{Text+tables+KB} setting (Table \ref{tab:kbc}).

\section{Analysis}\label{sec:analysis}

Having demonstrated that combining information sources does improve answer accuracy, we now provide more analysis on \emph{how} this is achieved
by inspecting both retriever and reader closely.

\paragraph{Retriever}

One natural assumption is that adding more data increases the coverage of relevant contexts that can be used to answer the input questions, thereby improving the end-to-end performance.
We verify this by examining the retrieval results of different models using the NQ development set, where a context is considered relevant if it contains the correct answer string. 
When more knowledge sources are added, our system is able to improve retrieval \emph{recall} (Table~\ref{tab:tables}, top half), which may correlate with the end-to-end answer accuracy shown in~Table \ref{tab:main}.

\begin{table}[t]
\begin{center}
\begin{tabular}{lrr}
\toprule
\textbf{Model} & \textbf{R@20} & \textbf{R@100} \\ 
\midrule
Text-only  & 80.0 & 85.9  \\
w/ lists & 82.7 & 89.6 \\ 
w/ tables & 83.1 & 91.0 \\
w/ lists + tables & 85.0 & 92.2 \\
w/ lists + tables + KB & 83.4 & 92.8 \\
\midrule
\multicolumn{3}{c}{\textit{Tables-only}}\\
\midrule
simple linearization & 86.3 & 94.3 \\
template linearization & 60.8 & 69.4 \\
\bottomrule
\end{tabular}
\end{center}
\caption{Retrieval recall on the NQ dev set with different settings. Tables only results are for the NQ dev subset which has answers in tables.}
\label{tab:tables}
\vspace{-3mm}
\end{table}

\paragraph{Reader}

Although including additional information sources improves the chance of retrieving relevant contexts, 
it is not guaranteed that the reader can leverage those contexts and output the correct answers.
For instance, reader model training may benefit from diverse sources of contexts, and the end-to-end improvement of answer accuracy may simply be attributed to a reader model that performs better on contexts from regular text.
Due to the nature of the FiD generative reader, however, it is non-trivial to ascertain which input context(s) contribute the answer.
As a proxy, we look at the correlation between the source of \emph{positive} contexts (those which contain a correct answer string) feeding into the reader model and the performance change in the outcome. 

\begin{figure}[t]
    \centering
    \includegraphics[width=\linewidth]{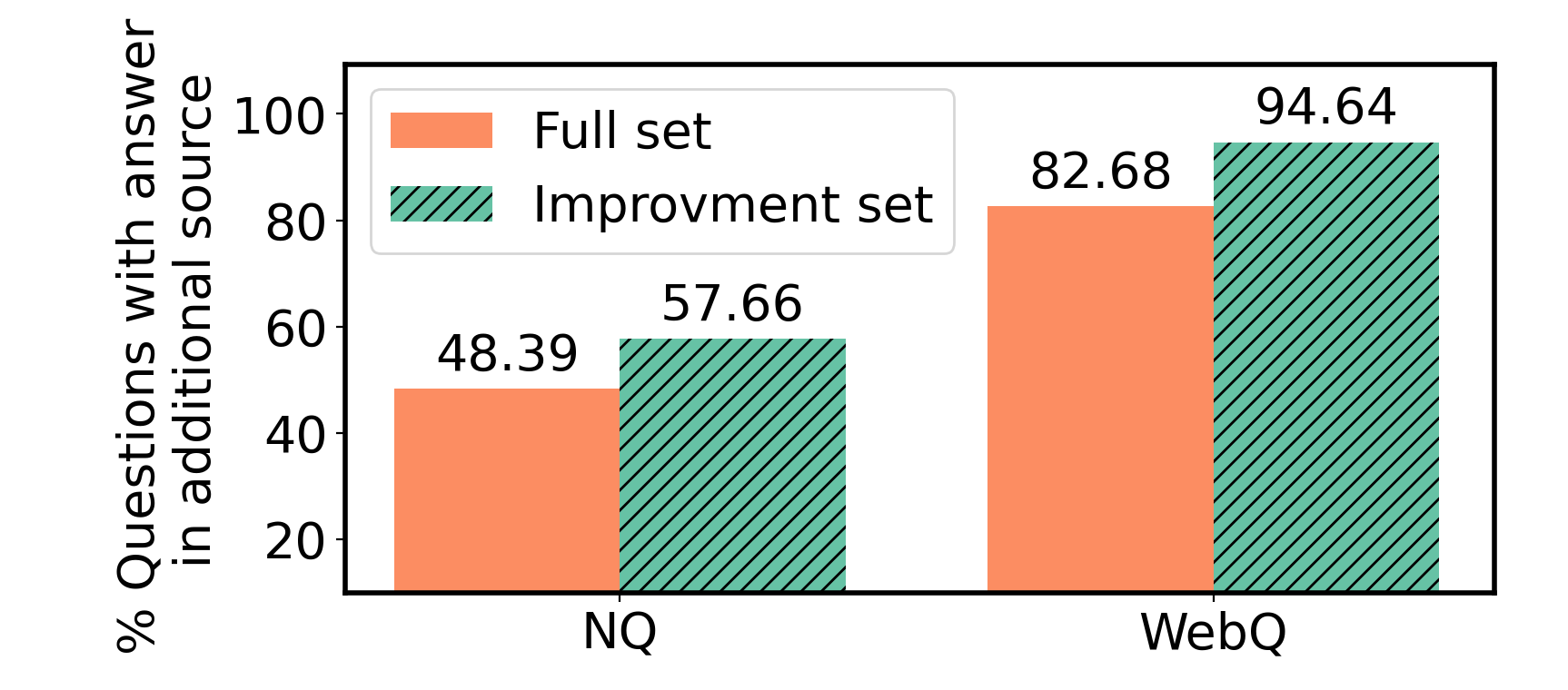}
    \caption{Percentage of questions with answers in additional sources. For NQ the additional sources are list and tables. For WebQ the additional source is KB.}
    \label{fig:source_improvement_barchart}
    \vspace{-3mm}
\end{figure}

Suppose we are comparing two reader models $M_u$ and $M_t$, where $M_u$ uses additional sources of information compared to $M_t$ (e.g., $M_t$ uses text only and $M_u$ uses text and KB).
Let $Q$ be all the questions in our development set, $Q_u \subseteq Q$ and $Q_t \subseteq Q$ the subsets of questions answered correctly by $M_u$ and $M_t$, respectively.
The \emph{improvement} set $Q' = Q_u - Q_t$ is thus the questions that $M_u$ manages to improve upon $M_t$.
Examining the source of the positive contexts for the questions in $Q'$ can help shed some light on how $M_u$ performs better.
For example, if more positive contexts are from KB rather than text, then the improvement is more likely due to additional information present at inference time.
Figure~\ref{fig:source_improvement_barchart} plots the percentages of positive contexts originating from the additional sources for the questions in the full development set ($Q$) vs those in the improvement set ($Q'$) in two cases. 
The first one compares a baseline \emph{text}-only model to a model with lists and tables added on NQ, and
the second compares a \emph{text+tables} model with \emph{text+tables+KB} on WebQ.  
In both cases, answers retrieved from the additional source correlate with a better outcome. 

To examine the effects of other indirect factors, such as the change of overall model quality due to the inclusion of varied sources or more training samples from the tables, we evaluate the \emph{text + tables} model with text-only input.  
We find that this achieves similar performance (48.7 EM) on the NQ test set compared to a \emph{text}-only model on the same input, suggesting that these other factors are not a major contributor and that the improved performance is primarily due to the added knowledge from structured sources.

\section{Implementation Details}\label{sec:implementation}

The code, data, and trained model checkpoints of \uniqa{} are available at: \url{https://github.com/facebookresearch/UniK-QA}.
\subsection{DPR Training}\label{sec:implementation:dpr}
Our DPR model is trained on the entire Wikipedia dump, including lists and tables, as described in~\secref{sec:model:tables}.
Specifically, lists are treated as normal texts and included in standard text passages, while tables are converted to their own ``passages'' using our linearization approach.
We combine all these passages from the text, lists and tables into the Wikipedia passage collection, and train DPR using the standard setup~\cite{DPR}: We use BERT-base \citep{BERT} encoders, 100-token text passages, and a single negative document per question.  Negatives are mined with BM25 in the first iteration, and from the first iteration model for the second iteration.  We train for 40 epochs with a linear warmup of 500 steps, batch size of 128 and learning rate $10^{-5}$.

As mentioned in \secref{sec:model:kbqa}, we do not retrain DPR for retrieving KB relations.
The public DPR checkpoint for open-domain question answering is used in our WebQSP experiment (\secref{sec:exp:kbqa_as_textqa}), while we use our own DPR model trained on text, lists and tables for retrieving KB relations in our multi-source QA experiments (\secref{sec:experiments}).

\begin{table*}[ht]
    \centering
    \begin{tabular}{l c c c c c}
    \toprule
    KB Quota && NQ & WebQ & TREC & Trivia \\
    \midrule
    Wikipedia + Wikidata && 10 & 30 & 10 & 10 \\
    Wikipedia + Freebase && 10 & 40 & 10 & 20 \\
    Wikipedia + Wikidata \& Freebase && 10 & 30 & 10 & 20 \\
    \bottomrule
    \end{tabular}
    \caption{The quota of ``passages'' converted from KB relations in each experiment.}
    \label{tab:kb_quota}
\end{table*}

\subsection{FiD Training}\label{sec:implementation:fid}
We adopt the FiD model with T5-large~\cite{t5} and 100 context documents and use the original hyper-parameters of FiD~\cite{izacard-grave-2021-leveraging} whenever possible.
In particular, the Adam~\cite{kingma2014adam} optimizer is used with a constant learning rate of $0.0001$.
The model is trained for $10k$ steps, with a batch size of 64, using 64 V100 GPUs.
We did not perform any hyper-parameter search.

\subsection{Merging KB and Text}\label{sec:implementation:merge}

As mentioned in \secref{sec:exp:combination}, we tune the quota for KB relations by maximizing retrieval recall on the development set.
Table~\ref{tab:kb_quota} shows the number of KB ``passages'' (out of 100 total context passages) selected in our final model.
The text and KB passages are interleaved in the final context passages.

For each dataset, the KB quota (which can also be interpreted as the helpfulness of the KB) is relatively stable across different choices of KBs.
WebQuestions has the highest KB quota, which is expected given that it was originally collected as a KBQA dataset.
Experimental results in Table~\ref{tab:main} also confirm that using KB brings the most gains on WebQuestions.

\section{Discussion}
We demonstrated a powerful new approach, \uniqa{}, for unifying structured and unstructured information sources for open-domain question answering.  We adopt the simple and general \emph{retriever-reader} framework and show not only that it works for structured sources, but improves over traditional KBQA approaches by a wide margin.  By combining sources in this way, we achieved new state-of-the-art results for two popular open-domain QA benchmarks.

However, our model also has several shortcomings in its current form.  As a result of flattening all sources into text, we lose some desirable features of structured knowledge bases:  the ability to return \emph{all} answers corresponding to a query, and the ability to infer multi-hop paths to answer more complex questions.  In this work we have side-stepped the first issue by focusing on the exact match metric (equivalent to  Hits@1), which is standard in the open-domain QA literature, but largely ignores multiple answers.  We were also able to ignore the second issue, since the datasets we evaluated on, while standard, are composed mostly of simple, natural user questions which can be answered from a single piece of information.

We do believe these are important details and they can be addressed within the framework described here. For instance, outgoing edges of an entity with the same relation can easily be merged, thus encoding all answer entities into a single text representation.  It is also possible to simply generate multiple answer candidates from the reader's decoder.  For multi-hop question answering, there is recent work \citep{xiong2020answering} successfully extending dense retrieval to the multi-hop setting \citep{yang-etal-2018-hotpotqa, welbl-etal-2018-constructing}, which could naturally be applied within our framework.  It remains to be seen how these approaches would compare to more traditional structured methods.

\bibliography{anthology,acl2021}

\end{document}